# Deep Artificial Intelligence for Fantasy Football Language Understanding

Aaron Baughman
IBM
Cary NC, USA
baaron@us.ibm.com

Micah Forester
IBM
Austin TX, USA
mforste@us.ibm.com

Jeff Powell
IBM
Atlanta GA, USA
jjpowell@us.ibm.com

Eduardo Morales
IBM
Coral Gables FL, USA
Eduardo.Morales@ibm.com

Shaun McPartlin
Disney ESPN
Bristol CT, USA
Shaun.McPartlin@disney.com

Daniel Bohm
Disney ESPN
Bristol CT, USA
Daniel.Bohm@disney.com

**ABSTRACT**

Fantasy sports allow fans to manage a team of their favorite athletes and compete with friends and fellow managers. The fantasy platform aligns the real-world statistical performance of athletes to fantasy scoring and has steadily risen in popularity to an estimated 9.1 million players per month spending a total of 7.7 billion-million minutes with 4.4 billion player card views on the ESPN Fantasy Football platform from 2018-2019. In parallel, the sports media community produces news stories, blogs, forum posts, tweets, videos, podcasts and opinion pieces that are both within and outside the context of fantasy sports. However, human fantasy football players cannot consume and summarize billions of bytes of natural language text and multimedia data to make a roster decision. Before our system, fantasy managers relied on expert projections and their analysis of, on average, 3.9 sources of information to make roster decisions. While these experts excel at evaluating players based on traditional statistics, they are omitting the majority of data that would inform their assessments.

Our work discusses and shows the results of a novel (patent pending) machine learning pipeline to effectively manage an ESPN Fantasy Football team. The use of trained statistical entity detectors and document2vector models applied to over 50,000 news sources and 2.3 million articles, videos and podcasts each day enables the system to comprehend natural language with an analogy test accuracy of 100% and keyword test accuracy of 80%. Next, deep learning feedforward neural networks that are 98 layers deep provide player classifications such as if a player will be a bust, boom, play with a hidden injury or play meaningful touches with a cumulative 72% accuracy with a 12% real world distribution. Finally, a multiple regression ensemble accepts the deep learning output and ESPN projection data to provide a point projection for each of the top 500+ fantasy football players in 2018. The point projection maintained a Root Mean Squared Error of 6.78 points. Next, the best out of the 24 probability density functions that were fit to the current projection and historical scores was selected to visualize score spreads. Within the first 6 weeks of the product launch, the total number of users spent a cumulative time of over 4.6 years viewing our AI insights. As a result, each user spent over 90 seconds using the evidence from our novel algorithms. The training data for our models was provided by a 2015 and 2016 web archive from Webhose, ESPN statistics, and Rotowire injury reports. We used 2017 fantasy football data as a test set.

## 1 Introduction

Fantasy sports owners and managers have hundreds of critical questions to answer before selecting their team. Who will score the most points this week? What player will be a bust or breakout? Will any players be a sleeper? Do any players have injuries that are going to impact their play? When should a player start to counter my opponent's team? Are there any available trades to upgrade a team? The number of possible moves to make is daunting for both the professional and novice player.

With over 9.1 million unique fantasy football players per month on the ESPN platform alone, the demand for content is insatiable. Every day during the 2018 season, we sustained 2 billion edge hits and delivered 250 TB of AI content per day. The large volume of users bases the majority of their roster decisions on player rankings and simple statistics. However, unstructured and multimedia information about sports is the largest data component. The volume of natural language, video, and podcast content creates fantasy football content shock that is an epidemic for current team managers.

The overwhelming majority of fantasy sports participants filter content based on personal biases such as reading articles, watching videos, or listening to podcasts about their favorite team or from their preferred outlet. On average, fantasy players consume 3.9 sources to base their decisions [1]. Other users rely on ad-hoc tools such as querying statistics databases, excel sheets, or natural language searches [1,2]. The limited amount of information each manager can consume has created tremendous knowledge gaps when making decisions.

Throughout the 2018 NFL football season, we developed a novel system that reads and comprehends natural language, videos, and podcasts from over 50,000 sources that were deployed to the ESPN Fantasy Football mobile and desktop experiences. The semantic relationships between words and topical understanding through techniques such as doc2vec enabled deep learning classifiers to make decisions about each football player. Team managers and coaches now have insight from unstructured textual and multimedia data as to which players will be a bust, breakout, play meaningful touches, or play with a hidden injury. Statistical data is combined with our system's comprehension of unstructured information through a deep machine learning pipeline. A best-fit score distribution from a set of 24 probability density functions (PDF) based on current and historical score projections provide an understandable score spread. To answer the question why, we presented the top 10 articles, podcasts, and videos that support or refute our machine learning pipeline's player assessment. This paper depicts the empirical evaluation of our ESPN Fantasy Football Insights with Watson system (FFIW). To protect trade secrets and to maintain our competitive advantage within fantasy sports, we will show a limited amount of user feedback.

## 2 Abbreviated Related Works

Advanced analytics that use structured data such as historical game statistics are prevalent and widely used by fantasy sports managers. For example, Rotogrinder provides a service that builds starting lineups, player projections, Vegas odds, depth charts, and weekly weather reports. Another tool available for fantasy sports players is called Dailyfantasynerd. The service highlights favorable statistics for players, maintains a lineup optimizer, and displays weather data for each venue. Fantasyfootballanalytics.net exposes aggregated play statistics for analysis along with custom point projections and player risk assessments.

Tools that ingest, consume and utilize a wide range of unstructured data such as text and multimedia have had limited utility for direct team management. On rotowire.com, fantasy sports managers can ask a human expert that has curated both structured and unstructured data for advice. ESPN has an insider paid service to access premium content written by featured columnists as well as roster advisors. SportsQ has a natural language question and answer system to retrieve passages relevant to a question. However, none of the prior work distills millions of articles, videos, and podcasts every hour into AI insights.

### 2.1 Machine Learnin in Fantasy Football

All of the prior within fantasy sports have been around statistics and structured data. Currently, we are not aware of any previous works that have analyzed multimedia and text information for the basis of computational guided fantasy sports play. For example, Landers and Duperrouzel present several machine learning approaches for predicting points scored by players as well as strategies to optimize a team. Other works use statistical predictors from sports play to optimize teams [4]. Following the general body of work within fantasy sports, Hermann et al. show how regression, naïve bayes, and decision trees can be used to predict fantasy basketball performance. Other works analyze predicting or projecting players' values within fantasy sports [6,7,8]. Seal addresses any doubt that machine learning with statistics can improve fantasy play. We further the state of the art and show how multimedia and natural language processing can provide quantitative and qualitative benefits to fantasy football experiences.

### 2.2 Deep Learning for Natural Langauge Processing

Large-scale text classification has been inspired by a growth of natural language text over social media and news outlets. Work by Glorot showed that stacked denoising auto-encoders performed text sentiment classification better than SVM's, Structural Corresprence Learning (SCL), Multi-label Consensus Training (MCT), and Spectral Feature Alignment [9]. Other deep learning works used convolutional neural networks and transfer learning by sharing network levels for auxiliary tasks to provide Semantic Role Labeling (SRL) [10].

Seminal work culminated by Tomas Mikolov in 2013 outlined the beginnings of the application of deep learning to natural language processing with doc2vec. Billions of words can be added to a computing system's vocabulary, which progresses the traditional n-gram language model [11]. Additional work shows that distributed word vector representations improve text

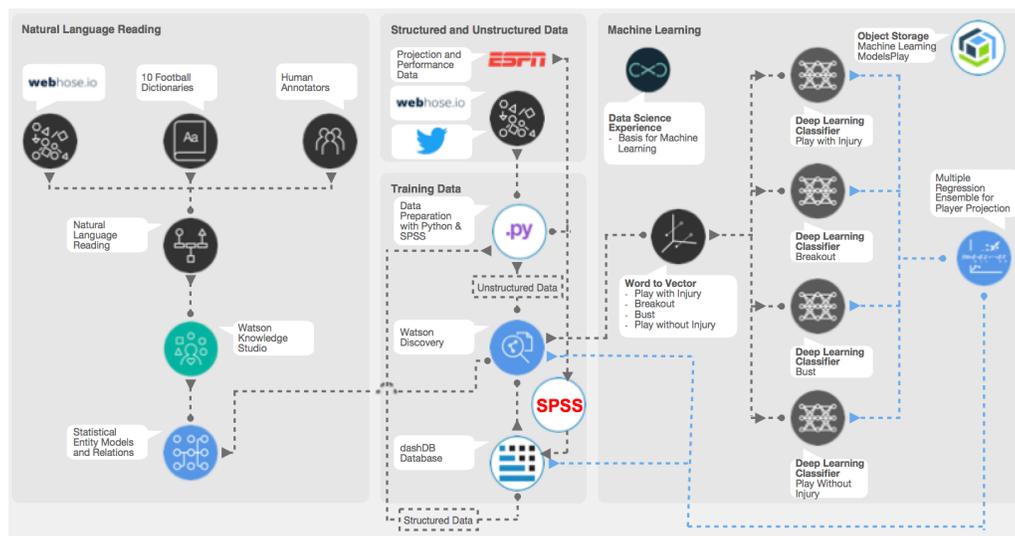

Figure 1: Machine Learning Architecture

classification over Bag of Words (BoW) and Support Vector Machines (SVM) [12].

Works began to use doc2vec approaches to expand short text for improvements in text classification [13]. Other works enriched short text using tf-idf measures before using a doc2vec approach for text classification [14]. Text embeddings were combined with multimedia convolutions within multitask learning to improve model performance [15]. Word2vec approaches are used within language translation to emphasize the use of unlabeled data between monolingual data [16]. Unlike the previous works, we extend doc2vec by summarizing thousands of documents into entities, concepts, and keywords before creating average word embeddings.

## 3 Machine Learning Pipeline

The machine learning architecture of the system is composed of five applications, dozens of models, several data sources, and data science environments. Figure 1 depicts the overall data flow within our system. First, the system had to be taught to read fantasy football content. A novel language model was designed with custom entities and relationships to fit the unique language people use to describe players and teams in the fantasy football domain. Next, an annotation tool called Watson Knowledge Studio was used by 3 human annotators to label text within articles as any combination of 13 entity types such as player, team, performance, etc. With this data, a statistical entity detector was trained and deployed to our system called Watson Discovery (WD) that continually ingests sources from over 50,000 sources. Podcasts and videos are transcribed and ingested into WD. The WD system is able to discover fantasy football entities, keywords, and concepts from the continually updating corpora based on our trained statistical entity model.

Next, the system used a document to vector model to understand the natural text from a query. A very specific query was initially issued to WD such as "Tom Brady and Patriots and NFL and Football." If a query did not return at least 50 documents, experimentally determined, the query was broadened until it only had "Tom Brady and NFL." From the query result, a list of entities, keywords and concepts for each document was converted to numerical feature vectors. Each of the feature vector groups (i.e. entities) was averaged together to represent a semantic summarization. All of the feature vector groups from each document were averaged across documents. The 3 feature vectors, along with player biographic data were input into the deep learning portion of the pipeline.

The deep learning pipeline phase had 4 models that were over 98 layers deep. The models were classifiers for each player to determine the probability of a boom, bust, play with a hidden injury or play meaningful minutes. The probability scores provide a confidence level of player states so that team owners can decide their own risk tolerance.

Finally, the outputs of the deep learning layers, along with structured ESPN data were input into an ensemble of multiple regression models. This merging of natural language evidence with traditional statistics produced a score projection for every player. On average, the combination of structured and unstructured data produced a better RMSE than each independently. Finally, 24 PDF's were fit to the score projection and historical score trends to produce a player score distribution. A lot of the data exploration was performed within Jupyter notebooks and SPSS. Through experimentation, we selected model hyperparameters and algorithms.

### 3.1 Fantasy Football Training Data

Throughout the project, we used historical news articles, blogs, etc. associated with players from the fantasy football seasons 2015 and 2016. A third party named Webhose provided the large-scale content. In total, over 100 GBs of data was ingested into WD using our custom entity model. We correlated the article date with structured player data from ESPN to generate labeled data. The ESPN player data contained several statistics that included week result, projection, actual, percentage owned, etc.

A week span date from Tuesday to the following Monday was associated with each player state so that a time ranged query could be run to retrieve relevant news articles. Equations 1-3 depict the determination of a boom label. If the actual score of a player was greater than 1 standard deviation above the projection for a player $p$, the weighted average of the differences between the actual and projected score by the percentage owned, $perowned_p^{0.1}$, is used to determine a boom standard deviation. However, the player must be owned by at least 10% in all leagues.

$$\mu_{bo} = \frac{1}{N}\sum_{p=0}^{N}\frac{(actual_p - projected_p)}{perowned_p^{0.1}}; actual_p > (projected_p + \sigma_p) \quad (1)$$

$$\sigma_{bo}^2 = \frac{1}{N}\sum_{p=0}^{N}\left(\frac{(actual_p - projected_p)}{perowned_p^{0.1}} - \mu_{bo}\right)^2; actual_p > (projected_p + \sigma_p) \quad (2)$$

The standard deviation for boom, $\sigma_{bo}$, is determined by taking the square root of the boom variance, $\sigma_{bo}^2$.

The label boom is applied to the player if their actual score, $x$, is greater than 1 boom standard deviation above the boom mean for the player.

$$boom(x) = \begin{Bmatrix} 1: x \geq \mu_{bo} + \sigma_{bo} \\ 0: x < \mu_{bo} + \sigma_{bo} \end{Bmatrix} \quad (3)$$

The bust label is calculated by equations 4-6. The average bust score, $\mu_{bu}$, is determined by weighting the difference between score actuals and projection by the same player's projection. However, only actuals that are 1 standard deviation, $\sigma_p$, below the projected scores are used within the sample set.

$$\mu_{bu} = \frac{1}{N}\sum_{p=0}^{N}\frac{(actual_p - projected_p)}{projection_p} * \sqrt{projected_p}; actual_p < (projected_p - \sigma_p) \quad (4)$$

$$\sigma_{bu}^2 = \frac{1}{N}\sum_{p=0}^{N}\left(\left(\frac{(actual_p - projected_p)}{projection_p} * \sqrt{projected_p}\right) - \mu_{bu}\right); actual_p < (projected_p - \sigma_p) \quad (5)$$

The square root of the bust variance, $\sigma_{bu}^2$, provides the standard deviation threshold to label a player with score $x$.

$$bust(x) = \begin{Bmatrix} 1: x \geq \mu_{bu} + \sigma_{bo} \\ 0: x < \mu_{bu} + \sigma_{bo} \end{Bmatrix} \quad (6)$$

The play with injury label was generated only for players that scored greater than 15% of their projected points and they were on the Rotowire injury report as questionable or probable. The play meaningful minutes label was created when a player scored greater than 15% of their projected points and was probable or not on the Rotowire injury report. Each of the four labels was generated for every week of every player within the fantasy football 2015 and 2016 seasons.

### 3.2 Statistical Entity Detection

The system had to learn how to read fantasy football documents, blogs, and news articles and listen to videos and podcasts. In order to read text and transcripts for comprehension, an ontology of 13 entity types were defined that covered player-centric understanding. The entities include body part, coach, fans, gear, injury, location, player, player status, treatment, positive tone, negative tone, team and performance metric. 1,200 documents created a representative distribution of entities that were comparable to the 5,568,714 training and test document set.

A team of 3 human annotators used a tool called Watson Knowledge Studio to annotate text as 1 of 13 entity types. The documents were pre-annotated from 10 dictionaries that searched for words and automatically created an initial annotation. The annotators corrected pre-annotations while adding others that were missed. Each day, the team met to discuss their kappa statistic or agreement score between each other over each entity type. Over a span of 3 weeks, the team produced a statistical entity model with a precision of 79%, recall of 73% and an F1 score of 76%. Even with a 14% entity word density over all documents, the overall annotator agreement score was at 70% with the majority of differences being the omission of a few words in a phrase.

### 3.3 Document2Vector

For each queried document, a summarization of keywords, concepts and entities from the document is unioned into feature vector $s_i$ from Equation 7. A doc2vec model was trained on 50 documents for each day of the top 300 players from each of the previous Fantasy Football seasons. In total, the training set included 94 GB of text. A second precise oriented doc2vec model was trained on 10 football dictionaries. In total, both models learned millions of vocabulary words during the training stage. The feature vector is input into both of the broad word embedding model and narrow encyclopedia model for a spatial word embedding summarization of the document. As shown in Equation 8, the average spatial meaning of all the documents for a particular player from the encyclopedia embedding, $w_e$, and the broad embedding, $w_b$.

$$\overline{s_i} = \overline{k_d} \cup \overline{c_d} \cup \overline{e_d} \qquad (7)$$
$$\overline{a_p} = \frac{1}{N}\sum_{l=0}^{N} w_e(s_l) \cup w_b(s_l) \qquad (8)$$

### 3.4 Deep Player Classification

Every player was probabilistically classified as a weekly boom, bust, play with hidden injury, or play meaningful touches. Each player state used an identical neural network topology. The topology includes 98 layers where 6 sequential paths are merged into a series of densely connected neurons. Every parallel sequence leveraged a dropout layer at the beginning to prevent overfitting the gradients. With the large number of layers, batch normalization was used in each parallel path to speed up training.

Through experimentation, the play with hidden injury and bust classifiers used the tanh activation function from Equation 9 while the play meaningful touches and breakout classifiers implemented the relu activation function from Equation 10. The last layer uses a sigmoid activation function shown in Equation 11 to scale the output between 0 and 1. The networks used stochastic gradient decent to minimize the binary cross entropy. For each of the measurable player states such as boom and bust, a deep neural network has been designed and trained, $dnn_{state}$. Equation 12 shows the union of feature vectors average word embedding, $\overline{a_p}$, player bio vector, $\overline{b_p}$, and the social sentiment measures of a player, $\overline{s_p}$, are fed forward into each deep neural network for a probabilistic measurement.

$$a_{tanh}(z) = \frac{1-e^{-2(z)}}{1+e^{-2(z)}} \qquad (9)$$
$$a_{relu}(z) = log(1+e^x) \qquad (10)$$
$$a_{sigmoid}(z) = \frac{e^z}{e^z+1} \qquad (11)$$
$$dnn_{state}(\overline{a_p} \cup \overline{b_p} \cup \overline{s_p}) = l_{96}\left(l_{95}\left(l_{...}\left(l_0(\overline{a_p} \cup \overline{b_p} \cup \overline{s_p})\right)\right)\right) \qquad (12)$$

The inputs into the neural network include the player biographic, word 2 vector outputs, and social sentiment of each entity type in the document set. The output of each neural network learned the relationships between the input vectors.

### 3.5 Player Score Probability Distribution

A multiple regression ensemble based on player position provided a point projection for each player. Equation 13 shows the general linear regression used for each position.

$$s(\overline{x}) = \beta_0 + \beta_1 x_1 + \cdots + \beta_{|x|} x_{|x|} \qquad (13)$$
$$pdf_j(h_o \ldots h_n \cup s_0 \ldots s_n) = min_{loss}(pdf_0, pdf_1, \ldots, pdf_{25}) \qquad (14)$$

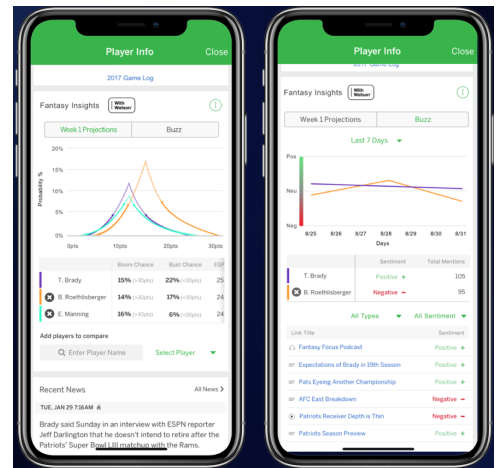

Figure 2: Mobile application player compare experience

Figure 2 shows a user interface that we created for Fantasy Football players. To produce the best PDF, the end of the machine learning pipeline fit 24 different PDFs to a player's historical and predicted performance as shown in Equation 14. Some example distributions include alpha, anglit, beta, bradford, chi, wald, vonmises, normal, rayleigh, etc. If the player did not have enough historical data, similar player data was retrieved.

The distribution that fit the data the best was selected to run 1,000 simulations or random draws. The simulations produced more likely real-world curves for player performances. We highlighted the 85th and 15th percentile on the graph so that users could easily compare players.

## 4 Results

Overall, the system provided informative and accurate Fantasy Football insights from text, video, audio, and statistics. The system projected 88.2% of players to be within 10 points of their projection and 71% of player scores to be within 7 points of a projection. From a score distribution perspective, 83% of players are within the high score range while 71% of players are within the low score range. Impressively, 90% of players either boomed or were close to boom when predicted to boom. On the other end, 78% of players either busted or were close to a bust when predicted to bust.

### 4.1 Document2Vector Analysis

The model was tested with two different types of semantic meaning evaluations. First, an analogy test was provided to the model. If the relation Travis Kelce is to the chiefs as Todd Gurley is to the X is presented to the model, the correct answer for X should be the Rams. In the player to team analogy testing, the correct answer was in the top 1% of the data 100% of the time. The team to location analogy was slightly lower, with a 93.48% accuracy because the natural queries were not focused around teams. The second test provided a set of keywords to the model and expected a related word. For example, if Tom Brady input into the model, we would expect to see the Patriots as output.

| Test | Subject | Criteria | Accuracy |
|---|---|---|---|
| Analogy | Players:Team | Top 500 (<1% of the data) | 100% |
| Analogy | Team:Location | Top 500 (<1% of data) | 93.48% |
| Keyword | Players | Top 70 | 80% |
| Keyword | Team & Location | Top 500 | 74% |

Table 1: Document2Vector Tests and Results

### 4.2 Deep Learning Results

The bust game classifier had an accuracy of 55% with a modest class separation, while the boom classifier had an accuracy of 67%. The bust classifier was optimized on real world player bust distribution and accuracy because players with high bust probabilities significantly overscored their projections on average. The bust players that were missed and marked incorrect were very close to the binary threshold of 0.5. Further, the negative predictive value of the bust model is 85.5% accurate and it produces a real-world percentage of bust players at 12%. Over predicting busts could be worse than a high accuracy. The accuracy number is not as meaningful an evaluation metric as the negative predictive value and percentage of players predicted to be a bust.

The play with injury classifier had an accuracy of 77% with a positive predictive value of 68.1%. The positive predictive value is very important for this classifier so that we know if a player is going to play with a hidden injury. The play meaningful minutes model produced an accuracy of 91.4%. The output of the class and probability provide valuable predictors for the score projections as well as insights about each football player.

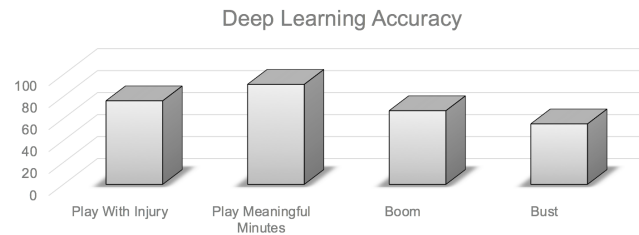

Figure 3: Deep Learning Results

From a real-world distribution of players that boom or bust, we were close to our objectives. Between 12-16% players generally boom while 30% can bust week over week. Figure 4 shows our results. Fantasy football users would quickly lose confidence in our system if we over predicted boom or bust.

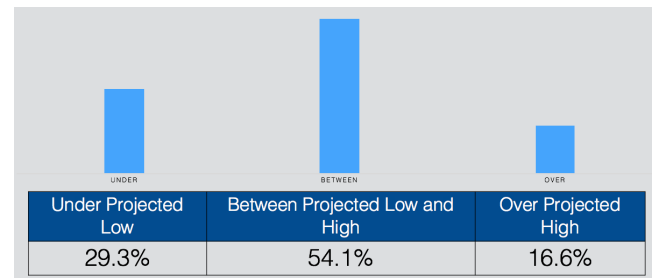

| Under Projected Low | Between Projected Low and High | Over Projected High |
|---|---|---|
| 29.3% | 54.1% | 16.6% |

Figure 4: Boom and Bust Player Distribution

### 4.3 Score Projection Results

A linear combination of deep learning player states and ESPN statistical data produced the best RMSE score of 6.78. On average, each player projected to score significant points over all positions will have a projection score that is off by 6.78 points when ESPN and our system is combined.

| Model | RMSE (Point Error) |
|---|---|
| ESPN Projection | 6.81 |
| Watson Adjusted Projection | 6.92 |
| Combined Projection | 6.78 |

Table 2: Point Projection Performance

The accumulation of the score projections over the duration of the football season provide data points for curve fitting as discussed in section 3.5. The low RMSE validates the probability distribution curve fits so that users can compare player shapes to each other as shown in Figure 2.

## 5 Community Impact

Each week throughout 2018, fantasy football team owners had the option of using our system to set team lineups. This was presented to fans in a few different forums such as player screens in the ESPN Fantasy App with segments aired on TV and in digital content. We found that empirical based decisions supported by the system helps to minimize the temptation of starting a player just because he is on your favorite team or is one of your favorite players. In September alone, over 5.5 billion insights were produced for the 9.8 million users that accessed the ESPN Fantasy App, and Watson complemented with evidence the 2.4 billion minutes fans spent in the app that month. This unprecedented level of depth and insight from unstructured data complimented by ESPN's traditional player statistics and analysis provided a comprehensive and detailed story about each player. Watson used some of that content as evidence to explain deep insights.

Towards the end of the 2018 ESPN Fantasy Football season, over a thousand players participated in a survey to measure the impact of our system. From the active survey respondents, over 80% of the users who utilized the feature said the Watson AI insights generated from our system helped them to enjoy fantasy football better. The more a fan followed the NFL, the more likely they used our system.

Some notable feedback included desire for a consistency metric, clear definition of boom/bust, simplified point potential, cumulative boom occurrences, and additional transparency around the models. Other users wanted the ability to compare more players that are and are not on a specific roster. These are all items we will be looking at as we continuously improve the experience.

From a marketing perspective, we had former NFL players, ESPN on-air talent, an IBM data scientist and movie star compete in a public ESPN influencer league. The influencer league and interest around our system generated 17.5 million impressions, 5.3 million video views, and 372 thousand total engagements from over 300 media pieces. The conversation on social media about ESPN Fantasy Football grew 24% in positivity from 2017.

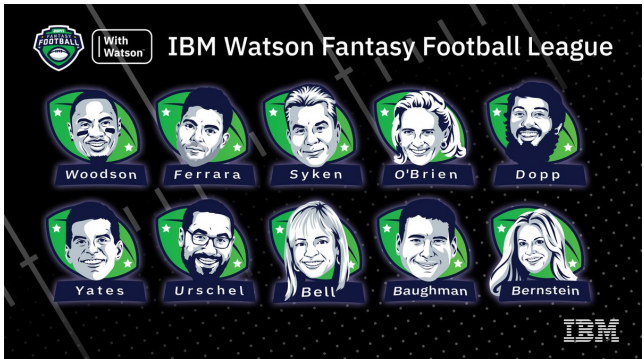

Figure 5: Fantasy Football Social Influencer League

| Source | Link |
|---|---|
| Explainer Video | https://youtu.be/xCszkWFAxmA |
| IBM Homepage | https://www.ibm.com/sports/fantasy |
| ESPN Fantasy Show | https://youtu.be/4BsDzKvBb3E |
| Blog Series | https://developer.ibm.com/series/watson-behind-the-code-fantasy-football-2018/ |
| Front of Code Commercial | https://www.youtube.com/watch?v=1cYrk67I00E |
| Podcast | https://bit.ly/2BQ3PS6 |

Table 3: Additional Impact Information

## 6 Future Work

To further our research and user experience using multimedia data throughout fantasy football, we are examining the possibility of developing a trade optimizer. The trade optimizer would find possible trading partners within a league and suggest a trade. We would like to examine the tradeoff between the likelihood of a trade being accepted with the utility gained by the initiating team. Our goals are to increase successful trades throughout fantasy football that help all teams involved within a transaction to increase their chance of winning. The experience will be engaging and insightful across mobile applications.

### ACKNOWLEDGMENTS
We would like to thank ESPN for their support by opening their television studios to our project. In addition, IBM Marketing, including Noah Syken, John Kent, Elizabeth O'Brien, and Kristi Kolski provided continuous feedback and encouragement. From a user interface perspective, we thank Jeffery Gottwald and Gary Guerno for their usability insights. At the beginning of the project, Rob High, an IBM Fellow, helped to make this project a reality. We appreciate paper reviews by Eduardo Morales. Finally, we thank Stephania Bell, Field Yates, Daniel Dopp, Matthew Berry, Charles Woodson, Bonnie Bernstein, John Urschel and Jerry Ferarra for using our platform through the 2018 ESPN Fantasy Football season.

### REFERENCES
[1] S. Hirsh, C. Anderson, and M. Caselli, "The Reality of Fantasy: Uncovering Information-Seeking Behaviors and Needs in online Fantasy Sports," in *Proc. CHI*, Austin Texas, May 5-10, 2012, ACM 978-1-4503-1016-1.

[2] G. Dzodom and F. Shipman, "Data-Driven Web Entertainment: The Data Collection and Analysis Practices of Fantasy Sports Players," in *Proc. WebSci*, Bloomington, IN, June 23-26, 2014, ACM 978-1-4503-2622-3/14/06.

[3] Landers, J. and B. Duperrouzel, "Machine Learning Approaches to Competing in Fantasy Leagues for the NFL." IEEE Transactions on Games, vol. 11, issue 2, 2019.

[4] T. Matthews, S. D. Ramchurn, and G. Chalkiadakis, "Competing with humans at fantasy football: Team formation in large partially-observable domains," in *Proc. AAAI*, 2012, pp. 1394–1400.

[5] Hermann, E and N. Adebia, "Machine Learning Applications in Fantasy Basketball.", semantic scholar, 2015.

[6] Seal, C., "Can machine learning help improve your fantasy football draft?", https://medium.com/fantasy-outliers/can-machine-learning-can-help-improve-your-fantasy-football-draft-4ceea1f1b2bd, 07/07/2019.

[7] N. Dunnington, "Fantasy football projection analysis," Ph.D. dissertation, Depart. Econ., Univ. Oregon, Eugene, OR, USA, 2015.

[8] R. Lutz, "Fantasy Football prediction," arXiv:1505.06918, 2015.

[9] X. Glorot, A. Bordes, Y. Bengio, "Domain Adaptation for Large-Scale Sentiment Classification: A Deep Learning Approach," in Proc. *ACM ICML*, 2011, pp. 513-520.

[10] R. Collobert and J. Weston, "A Unified Architecture for Natural Language Processing: Deep Neural Networks with Multitask Learning," in Proc 25[th] International Conference on Machine Learning, Helsinki, Finland, 2008.

[11] T. Mikolov, K. Chen, G. Corrado, and J. Dean, "Efficient Estimation of Word Representations in Vector Space," arXiv preprint arXiv:1302.3781, 01/16/2013.

[12] T. Mikolov, I. Sutskever, K. Chen, G. Corrado, J. Dean, "Distributed Representations of Words and Phrases and their Compositionality," Advances in neural information processing systems, pp. 3111-3119, 2013.

[13] P. Wang, B. Xu, J. Xu, G. Tian, C. Liu, H. Hao, "Semantic expansion using word embedding clustering and convolutional neural network for improving short text classification," Neurocomputing, Vol. 174, Part B, 01/22/2016, pp. 806-814.

[14] D. Yao, J. Bi, J. Huang, and J. Zhu, "A word distributed representation based framework for large-scale short text classification," in *IEEE IJCNN*, 2015, pp. 1-7.

[15] L. Kaiser, A. Gomez, N. Shazeer, A. Vaswani, N. Parmar, L. Jones, and J. Uszkoreit, "One Model To Learn Them All," arXiv:1706.05137v1 June 16, 2017.

[16] S. Jansen, "Word and Phrase Translation with word2vec," arXiv:1705.03127, 2017.